\documentclass {article}
\usepackage [latin1]{inputenc}
\usepackage{amsmath, amsthm, amssymb}[]
\usepackage{dsfont}
\usepackage{bm}
\newcommand{\funi}{\footnotesize\textbf{i\hspace{-0.22
cm}\normalsize$\bigcirc$
}}

\newcommand{\funciu}{\footnotesize\textbf{i\hspace{-0.22
cm}\normalsize$\bigcirc_{u}$}
}

\begin{document}
\title{Decision-making processes in \\ the Cognitive Theory of True Conditions}
\author{Sergio Miguel Tom\'e\\ \\
\small{Grupo de Investigaci\'on en Miner\'ia de Datos (MiDa)},\\
\small{Universidad de Salamanca, Salamanca, Spain}\\
\small{sergiom@usal.es}}

\maketitle
\begin{abstract}
The Cognitive Theory of True Conditions (CTTC) is a proposal to design the implementation of cognitive abilities and to describe the model-theoretic semantics of symbolic cognitive architectures. The CTTC is formulated mathematically using the multi-optional many-sorted past present future(MMPPF) structures.  This article discussed how decision-making processes are described in the CTTC.
\end{abstract}

\section{Introduction}
Semantics is one of the most challenging aspects of cognitive architectures. The Cognitive Theory of True Conditions (CTTC) is a proposal to implement cognitive abilities based on model-theoretic semantics and to describe the model-theoretic semantics of symbolic cognitive architectures \cite{Sergio2006,SergioTesis}.
The fundamental principle of the CTTC is that the perceptual space is a set of formal languages that denote elements of a model that is a quotient space of the physical space. At this moment, the mathematical formulation of the CTTC is using as model the \emph{multi-optional many-sorted past present future}(MMPPF) structures \cite{Sergio2018}. Also, the CTTC proposes a hierarchy of three formal languages to describe them.

This article discusses how decision-making processes are described in the CTTC. The article  is divided in two sections. The  first section addresses how the CTTC gives a functional equation that relates a MMPPF structure with the hierarchy of formal languages.  The second section addresses a set of solutions to generate behavior denominated heuristics of qualitative semantics, and its basis is detailed.

\section{The relation between the MMPPF structures  }

An important issue within cognitive science is how cognitive behaviour is generated. The CTTC addresses the issue  relating behaviour with a decision-making processes. Mathematically, the CTTC establishes a equation that relates the MMPPF structures with the hierarchic of languages.   The general equation is the following:

\begin{align}
\funciu(\bm{t})=\mathcal{I}^{\bm{r}_{p}}( \pi^{1} ( \pi^{1} ( F_{u}(\langle\phi_{1},...,\phi_{n}\rangle_{t},\langle\psi_{1},...,\psi_{n'}\rangle_{t-1} )  )  ))
\label{equ:general}
\end{align}

where, $F_{u}$ is a function that generates a pair of elements (the first element is a sequence of constants that denote interactions of the agent $u$, and the second is a tupla of formulas of the formal languages of the hierarchy),  $\langle\phi_{1},...,\phi_{n}\rangle_{t}$ is a tuple of formulas of the formal languages of the hierarchy that is input, $\langle\psi_{1},...,\psi_{n'}\rangle_{t-1}$  is a tuple of formulas of the formal languages of the hierarchy that was generated by the system, $\pi^{n}$ is the projection function that projects the n-element of a tuple, and  $\bm{r}_{p}= (\bm{t},\bm{\varepsilon},||, \bm{e}_{x})$ is the existing reality of the moment of time $\bm{t}$, $\mathcal{I}^{\bm{r}_{p}}$ an interpretation function, and \; $\funi\!_{u}$ is the interaction function of the agent $u$. With \ref{equ:general} is associated the following equation:

\begin{align}
\langle\psi_{1},...,\psi_{n'}\rangle_{t-1} = \pi^{2}( F_{u}(\langle\phi_{1},...,\phi_{n}\rangle_{t-1},\langle\psi_{1},...,\psi_{n'}\rangle_{t-2} )  )
\end{align}

Thus, the left-hand side of the equation are elements of the mathematical structure, and its right-hand side are elements of the hierarchy of formal languages.

The function $F_{u}$ belongs to a space of functions $\mathbb{F}_{u}$. The elements of the space of functions $\mathbb{F}_{u}$ can be described with  lambda expressions that we can reduce it to a normal form which is a sequence of constants that denote actions of an agent. Those expressions are simply typed lambda calculus where the ground types  are $\mathcal{G} = \{ LP_{MMPFM}, LP^{*}_{MMPFM}, CL_{MMPFM}, I_{u} \}$. The reason to use the framework of lambda calculus is because $F_{u}$ can be composed of functions of functions.

One of the possible equations that can be derived from the general equation is the following:

\begin{align}
\lambda time.\;\;\funciu(\bm{t})  =  \lambda term.\mathcal{I}^{\bm{r}_{p}}( ( \lambda\phi_{1},\phi_{2}.\vdash_{u}(\varphi,\varphi') ))
\label{equ:particular}
\end{align}

or in a traditional style:
\begin{align}
\funciu(\bm{t}) =  \mathcal{I}^{\bm{r}_{p}}(   \vdash_{u}(\varphi,\varphi'))
\label{equ:partclassic}
\end{align}

where $\varphi$ is a description about $\bm{r}_{p}$ made by the agent $u$, $\varphi'$ is a description about a future and existing reality, $\vdash_{u}$ is a function that determines a representation of an interaction that denotes an action of the agent $u$ to arrive from $\varphi$ to $\varphi'$, and it is defined in the following way:
\[
\begin{array}{cccc}
  \vdash_{u}: & LP.M_{(\varepsilon,||)} \times LP.M_{(\varepsilon,|\downarrow)}  &\longrightarrow & I_{u} \\
   & (\varphi,\varphi') & \mapsto \vec{a} &
\end{array}
\]
\[
LP.M_{(\varepsilon,||)}, LP.M_{(\varepsilon,|\downarrow)} \subset LP
\]
Equations \ref{equ:particular} and  \ref{equ:partclassic} express  that the action that the agent $u$ carries out at each time is the interpretation of the representation of the action calculated by $\vdash_{u}$. It must be noted that $\pi^{1}$ has been eliminated in \ref{equ:particular} and  \ref{equ:partclassic} because the sequence generated by $\vdash_{u}$ has only one element.

The relation formulated can be seen as a functional equation if $\vdash_{u}$ is considered an unknown variable. The problem is that we do not know a method to determine $\vdash_{u}$. The CTTC proposes that a set of solutions to the equation \ref{equ:partclassic} is the heuristics of qualitative semantics. In the following subsection, the basis of the heuristics of qualitative semantics is detailed.

Another example of equation that can be derived from the general equation is the following equation:

\begin{align}
\funciu(\bm{t}) = \mathcal{I}^{\bm{r}_{p}}(\pi^{1}(
  \circledast(\psi_{u}, \vdash_{u})(\varphi,\varphi' )))
\label{equ:learning}
\end{align}

where $\psi_{u}$ is the log formula of the agent $u$.
\[
\begin{array}{cccc}
  \circledast & LP\times \mathbb{P} &\longrightarrow & \mathbb{P} \\
   & (\psi_{u},\vdash_{u}) & \mapsto & \vdash'_{u}
\end{array}
\]

Equation \ref{equ:learning} describes a behaviour with learning because $\vdash_{u}$ is transformed into $\vdash'_{u}$ considering $\psi_{u}$, and $\vdash'_{u}$ is the function that generates the sequence of constants that denote actions of the agent $u$.

\subsection{Heuristics of qualitative semantics}
The heuristics of qualitative semantics is a decision-making processes based on model-theoretic semantics that is a solution to the equation \ref{equ:partclassic}. The heuristics of qualitative semantics are based on six facts. The first fact is that the elements of some metainformation alphabets are elements defined by a condition between a pair of elements of a property. Thus, these metainformation alphabets could be seen as sets of relations, and each element of these metainformation alphabets can be seen as a binary relation between two elements of a property.

\[
R \subset V_{p} \times  V_{p}
\]

The second fact is that the actions can be considered relations. This is because, given that an action can be seen in a simplified view as

\[
a : V_{p} \longrightarrow  V_{p}
\]

and every n-ary function can be seen as a n+1-ary relation, an action can be defined as:

\[
a \subset V_{p} \times  V_{p}
\]

The third fact is that because $a$ always produces changes of the same qualitative kind, and  it can be seen as a relation, there is an $R$ for which the following is fulfilled:
\[
a \subset R
\]

The fourth fact is that a goal can be considered an element of binary relation. This is because it consists of a change of the current value $x$ for a value $y$, and these two values form a pair $(x,y)$.

The fifth fact is that if the elements of a metainformation alphabet are considered relations they are jointly exhaustive and pairwise disjoint. Thus, there is one, and only one, $R$ for which:
\[
(x,y) \in R
\]

The sixth fact is  that the distance between the actual state and the target state of a goal is greater than or equal to the distance that exists between the initial state and the state generated by an action.

Taking these facts into account, we can conclude that

\[
(x,y) \in  R \supset a
\]

The previous conclusion justifies that an agent can achieve the goal $(x,y)$ by selecting the action $a$ labeled with $R$. Although one could claim that $(x,y)$, $R$ and $a$ are syntactic elements, $(x,y)$ and $a$ denote elements of the MMPPF structure, and $R$ is defined by elements that also denote elements of the MMPPF structure. The symbols must be in agreement with what they denote in the structure. Only if the symbols do not denote elements of an structure, they are the only important thing. However, one can observe that for a biological system or robot, it is important that an agreement exists between what the symbols denote and how the symbols are related among them. For example, if the syntactic definition of an action increases the temperature, but the opposite occurs when the agent uses that action, then there is a serious problem for the system to behave adequately in the environment. The relation between the symbols and what they denote must be in agreement, and if they are not, the heuristic does not work.

The Means-Ends Analysis (MEA)  of the GPS \cite{Newell1959,Ernst1969} works the same as the heuristics of qualitative semantics. MEA was the first heuristic created in AI.  The elements of the metainformation alphabets used in the heuristics of qualitative semantics are equivalent to the differences in the GPS. However, because Newell, Simon, and their collaborators were always focused on designing a program solver, they only gave the differences a role of classifying operators in the tables of connections.  They never considered the model-theoretic semantics involved or that the differences could be elements of a language that allowed an agent to describe its environment. In fact, we can consider MEA an element of CTTC, but the heuristics of qualitative semantics involve model semantics issues that are not foreseen by MEA.

The Multifunctional Robot on Topological Notions (MROTN) programme researches the use of topological notions to represent qualitative spatial relations and applies these notions to develop multifunction robots capable of social interaction \cite{Sergio2013,Sergio2014,Sergio2015,Sergio2017}. The CTTC and the MROTN programme share one problem. Both deal with how one qualitative decision-making process can be carried out from the quantitative data registered by sensors.  The CTTC addresses this problem using the hierarchy of languages and heuristics of qualitative semantics, and these two concepts have been used to develop the heuristics of topological qualitative semantics  \cite{Sergio2014,Sergio2015}. This research has shown how qualitative navigation can be done through unknown and dynamic open spaces \cite{Sergio2017}.

\bibliographystyle{splncs03}
\bibliography{BibCTTC}

\end{document}